\documentclass[final,12pt,5p,twocolumn]{elsarticle}

\usepackage{graphicx}
\usepackage{multirow}
\usepackage{url} 
\usepackage{multicol}
\usepackage{booktabs}
\usepackage{caption}

\usepackage{bm}
\usepackage{bbm} 
\usepackage{tabularx}
\usepackage{mathrsfs}
\usepackage{tabulary}

\usepackage{subfigure}

\usepackage{makecell}
\usepackage{amsmath,amsfonts,amssymb}
\usepackage[ruled,linesnumbered]{algorithm2e}  

\begin{document}
 	
 \begin{frontmatter}
\title{Bi-Decoder Augmented Network for Neural Machine Translation} 

\author[1]{Boyuan Pan}
\ead{panby@zju.edu.cn}
\author[2]{Yazheng Yang}
\ead{yazheng\_yang@zju.edu.cn}
\author[2]{Zhou Zhao}
\ead{zhaozhou@zju.edu.cn}
\author[2]{Yueting Zhuang}
\ead{yzhuang@zju.edu.cn}
\author[1,3]{Deng Cai\corref{cor1}}
\ead{dengcai@cad.zju.edu.cn}

\address[1]{State Key Lab of CAD$\&$CG, Zhejiang University}
\address[2]{Computer Science and Technology, Zhejiang University}
\address[3]{Alibaba-Zhejiang University Joint Institute of Frontier Technologies}
\cortext[cor1]{Corresponding author}

 \begin{abstract}
 	Neural Machine Translation (NMT) has become a popular technology in recent years, and the encoder-decoder framework is the mainstream among all the methods. It's obvious that the quality of the semantic representations from encoding is very crucial and can significantly affect the performance of the model. However, existing unidirectional source-to-target architectures may hardly produce a language-independent representation of the text because they rely heavily on the specific relations of the given language pairs. To alleviate this problem, in this paper, we propose a novel Bi-Decoder Augmented Network (BiDAN) for the neural machine translation task. Besides the original decoder which generates the target language sequence, we add an auxiliary decoder to generate back the source language sequence at the training time. Since each decoder transforms the representations of the input text into its corresponding language, jointly training with two target ends can make the shared encoder has the potential to produce a language-independent semantic space. We conduct extensive experiments on several NMT benchmark datasets and the results demonstrate the effectiveness of our proposed approach.
 \end{abstract}


\end{frontmatter}

\section{Introduction}
\label{sec1}
The encoder-decoder framework has been widely used in the task of neural machine translation (NMT)~\citep{luong2015effective,he2017decoding} and gradually been adopted by industry in the past several years~\citep{zhou2016deep,wu2016google}. In such a framework, the encoder encodes a source language sentence $x = \{x_1, x_2, ..., x_m\}$ into a sequence of vectors $\{\mathbf{h}_1, \mathbf{h}_2, ..., \mathbf{h}_m\}$ where $m$ is the length of the input sentence. The decoder generates a target language sentence $y = \{y_1, y_2, ..., y_n\}$ word by word based on the source-side vector representations and the previously generated words, where $n$ is the length of the output sentence. To allow the NMT to decide which source words should take part in the predicting process of the next target words, the attention mechanism~\citep{luong2015effective,bahdanau2015neural} has been applied widely in training neural networks, making models to learn alignments between different modalities.

Recently, transforming the text into a language-independent semantic space has gained significant popularity and it is a coveted goal in the field of natural language processing. Many methods proposed to learn cross-lingual word embeddings by independently training the embeddings in different languages with monolingual corpora, and then learn a linear transformation that maps them to a shared space based on a bilingual dictionary~\citep{artetxe2016learning,smith2017offline}. In this way, the encoder is given language-independent word-level embeddings, and it only needs to learn how to compose them to build representations of larger phrases. However, those methods explored to learn the representations of the low layers (\emph{i.e.}, the layers closer to the input words, such as word embeddings), while higher layers (farther from the input) which focus on high-level semantic meanings (similar to findings in the computer vision community for image features) are usually ignored~\citep{belinkov2017neural,guo2018soft}. Moreover, as the cross-lingual word embeddings are initially trained on the corpus of each language, we have to retrain and update them when we meet a new language case, which is very time-consuming.

For the encoder-decoder structure, there is another problem to be solved. As the model first encodes the source sentence into a high-dimensional vector, then decodes them into a single target sentence, it is hard to understand the meanings of the text and interpret what is going on inside such a procedure~\citep{shi2016does}. In the reality when we are translating a sentence to another language, we first comprehend it and summarize it into a language-independent semantic space (sometimes even an image), and then use the target language to represent it~\citep{robinson2012becoming}. The quality of the semantic space has a direct correlation to the performance of the translation and is one of the core problems of the natural language understanding.

\begin{figure*}[t]
\begin{center}
	\includegraphics[width=1 \textwidth]{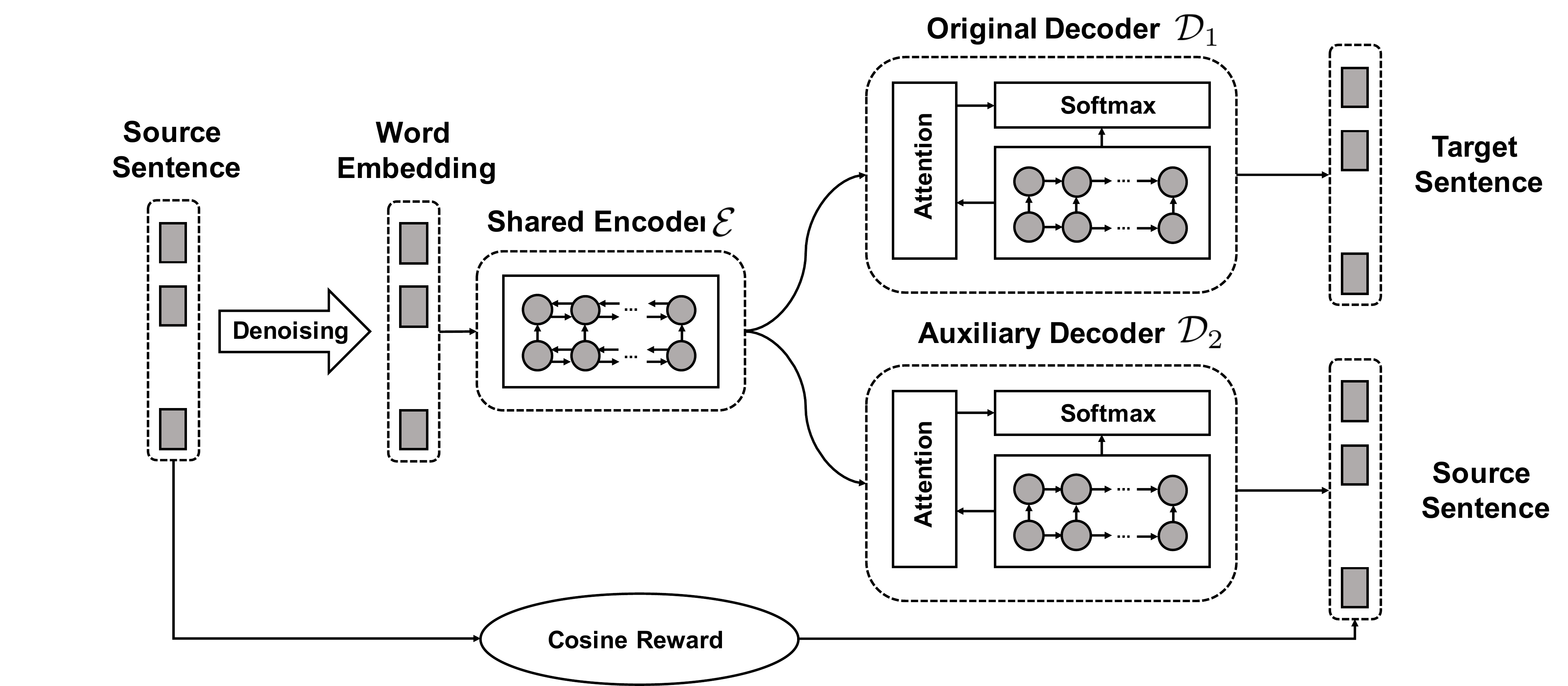}
	\caption{\label{fig1}Overview of the Bi-Decoder Augmented Network. For each input sentence of the source language, the original decoder and the auxiliary decoder output the sentence of the corresponding language based on the representations generated from the shared encoder.  }
\end{center}
\end{figure*}

In this paper, to avoid the tight dependency of the specific language pairs, we propose a novel Bi-Decoder Augmented Network for the task of the neural machine translation. Given the source input $x$, the encoder $\mathcal{E}$ firstly encodes it into high-level space vectors and the decoder $\mathcal{D}_1$ generates the target language sequences based on them. In addition to $\mathcal{D}_1$, we also design an auxiliary decoder $\mathcal{D}_2$, which composes an autoencoder with $\mathcal{E}$ to reconstruct the input sentence $x$ at the training time. By simultaneously optimizing the model from two linguistic perspectives, the shared encoder $\mathcal{E}$ could benefit from additional information embedded in a common semantic space across languages. At test time, we only use the well-trained $\mathcal{D}_1$ to output the target language. Moreover, because the reference sentence of $\mathcal{D}_2$ is the input sentence itself, we don't need any additional labor consumption or training data. To ensure that the training procedure of the bi-decoder framework captures the real knowledge of the language, we use strategies like reinforcement learning and denoising to alternately optimize different objective functions which we called \emph{multi-objective learning}. The contributions of this paper can be summarized as follows.

\begin{itemize}
\item Unlike the previous studies, we study the problem of neural machine translation from the viewpoint of the augmentation of the high-level semantic space. We propose a bi-decoder structure to help the model generate a language-independent representation of the text without any additional training data.

\item We incorporate the reinforcement learning and denoising for multi-objective learning in our training process to enable the autoencoder structure to learn the deep semantic knowledge of the language.

\item We conduct extensive experiments on several high-quality datasets to show that our method significantly improves the performance of the baselines.
\end{itemize}

The rest of this paper is organized as follows. In Section 2, we introduce our proposed bi-decoder framework. In Section 3, we show our training details of the model. A variety of experimental results are presented in Section 4. We provide a brief review of the related work about link prediction in Section 5. Finally, we provide some concluding remarks in Section 6.

\section{The Framework}
\label{sec2}
In this section, as shown in the Figure \ref{fig1}, we introduce the framework of our Bi-Decoder Augmented Network. The encoder-decoder models are typically implemented with a Recurrent Neural Network (RNN) based sequence-to-sequence structure. Such a structure directly models the probability $P(y|x)$ of a target sentence $y = \{y_1, y_2, ..., y_n\}$ conditioned on the source sentence $x = \{x_1, x_2, ..., x_m\}$, where $m$ and $n$ are the length of the sentence $x$ and $y$. For the auxiliary decoder $\mathcal{D}_2$, although the real target is still the source sentence $x$, we will uniformly denote it as $y$.

\subsection{Shared Encoder}
Given an input sentence $x$, the shared encoder $\mathcal{E}$ reads $x$ word by word and generates a hidden representation of each word: 

\begin{equation}
\label{eq1}
\begin{aligned}
\mathbf{h}_s = F_{enc}({\rm Emb}(x_s),\mathbf{h}_{s-1})
\end{aligned}
\end{equation}
where $F_{enc}$ is the recurrent unit such as Long Short-Term Memory (LSTM)~\citep{sutskever2014sequence} unit or Gated Recurrent Unit (GRU)~\citep{cho2014learning}, $x_s$ is the $s$-th word of the sentence x, ${\rm Emb}(x_s)$ is the word embedding vector of $x_s$, $\mathbf{h}_s$ is the hidden state. In this paper, we use the bi-directional LSTM as the recurrent unit. Compared to the word embeddings, the hidden states generated by the encoder are a higher-level semantic representation of the text, which is the source knowledge of the decoder. Therefore, a language-independent representation is very important to the comprehension of the text and directly corresponds to the quality of the neural machine translation.

\subsection{Bi-Decoder}
Initialized by the representations $\{\mathbf{h}_s\}_{s=1}^{m}$ obtained from the encoder, the decoders with an attention mechanism receive the word embedding of the previous word (while training, it is the previous word of the reference sentence; while testing, it is the previously generated word) at each step and generates the next word. 

Although the two decoders $\mathcal{D}_1$ and $\mathcal{D}_2$ output different languages and fulfill different responsibilities, their structures are the same, so we will use uniform symbols to denote them. Specifically, at step $t$, the decoder takes the previous hidden state generated by itself, previously decided word, and the source-side contextual vector as inputs. The hidden states of the decoder are computed via:

\begin{equation}
\begin{aligned}
\bar{\mathbf{h}}_t = F_{dec}({\rm Emb}(y_{t-1}),\bar{\mathbf{h}}_{t-1})
\end{aligned}
\end{equation}
where $F_{dec}$ is a unidirectional LSTM, $y_t$ is the $t$-th generated word, $\bar{\mathbf{h}}_t$ is the hidden state. For most attention mechanisms of the encoder-decoder models, the attention steps can be summarized by the equations below:

\begin{equation}
\label{eq3}
\begin{aligned}
\alpha_{ts} = \frac{{\rm exp(score}(\mathbf{h}_s,\bar{\mathbf{h}}_t))}{\sum_{s' = 1}^{m}{\rm exp(score}(\mathbf{h}_{s'},\bar{\mathbf{h}}_t))}
\end{aligned}
\end{equation}

\begin{equation}
\begin{aligned}
\mathbf{c}_t = \sum_{s} \alpha_{ts} \mathbf{h}_s
\end{aligned}
\end{equation}

\begin{equation}
\begin{aligned}
\mathbf{a}_t = g_a (\mathbf{c}_t, \bar{\mathbf{h}}_t) = {\rm tanh}(\mathbf{W}_a[\mathbf{c}_t; \bar{\mathbf{h}}_t] + \mathbf{b}_a)
\end{aligned}
\end{equation}
Here, $\mathbf{c}_t$ is the source-side context vector, the attention vector $\mathbf{a}_t$ is used to derive the softmax logit and loss, $\mathbf{W}_a$ and $\mathbf{b}_a$ are trainable parameters, the function $g_a$ can also take other forms. ${\rm score}()$ is referred as a \emph{content-based} function, usually implemented among different choices:

\begin{equation}
\begin{aligned}
{\rm score}(\mathbf{h}_s,\bar{\mathbf{h}}_t)=
\begin{cases}
\mathbf{W}[\mathbf{h}_s;\bar{\mathbf{h}}_t]\\
\mathbf{h}_s^{\top} \mathbf{W}\bar{\mathbf{h}}_t\\
\mathbf{v}^{\top}{\rm tanh}(\mathbf{W}_1\mathbf{h}_s + \mathbf{W}_2\bar{\mathbf{h}}_t)
\end{cases}
\end{aligned}
\end{equation}
where $\mathbf{W}, \mathbf{v}, \mathbf{W}_1, \mathbf{W}_2$ are trainable parameters, and we choose the third one as our content-based function based on the experimental results. The attention vector $\mathbf{a}_t$ is then fed through the softmax layer to produce the predictive distribution formulated as:

\begin{equation}
\label{eq13}
\begin{aligned}
P(y_t|y_{<t}, x) \propto {\rm softmax}(\mathbf{W}_p \mathbf{a}_t + \mathbf{b}_p)
\end{aligned}
\end{equation}
where $\mathbf{W}_p, \mathbf{b}_p$ are trainable parameters. We don't share the parameters of the attention mechanism between two decoders $\mathcal{D}_1$ and $\mathcal{D}_2$, and in this paper we regard them as a part of decoder.

\section{Training}
Denote $\theta_1$ as the parameters of the original decoder $\mathcal{D}_1$, $\theta_2$ as the parameters of the auxiliary decoder $\mathcal{D}_2$, $\theta_{e}$ as the parameters of shared encoder $\mathcal{E}$, ($D_{x}$, $D_{y}$) as the source-target language pairs of the training dataset. The training process of the encoder-decoder framework usually aims at seeking the optimal paramaters that encodes the source sequence and decodes a sentence as close as the reference target sentence. For the formula form, let $\Theta_{1} = [\theta_{e} ; \theta_1]$, the objective function for the decoder $\mathcal{D}_1$ is the maximum log likelihood estimation:

\begin{equation}
\begin{aligned}
\mathcal{J}_1 (\Theta_1) & =  \sum_{x \in D_x, y \in D_y} P(y|x; \Theta_1)\\
& = \sum_{x \in D_x, y \in D_y} \sum_{t = 1}^{n} {\rm log} P(y_t|y_{<t}, x; \Theta_1)
\end{aligned}
\end{equation}

Nevertheless, in our model we have two decoders so there are two different ends for the corresponding reference target sentences, which means that the final result of the optimization is a combination of two decoders. For the decoder $\mathcal{D}_2$, the most intuitive objective is similar to the above one: 

\begin{equation}
\begin{aligned}
\mathcal{J}_2 (\Theta_2) & =  \sum_{x \in D_x} P(x|x; \Theta_2)\\
& = \sum_{x \in D_x} \sum_{t = 1}^{m} {\rm log} P(x_t|x_{<t}, x; \Theta_2)
\end{aligned}
\end{equation}
where $\Theta_{2} = [\theta_{e} ; \theta_2]$. However, this objective function may not be the optimal one because the training procedure with the same input and output sentences essentially tends to be a trivial copy task. In this way, the learned strategy for the model would not need to capture any real knowledge of the languages involved, as there would be many degenerated solutions that blindly copy all the elements in the input sequence. To help the model learn the deep semantic knowledge of the language, we train our system using the following strategies.

\subsection{Denoising}
From the perspective of $\mathcal{D}_2$, the model takes an input sentence $x$ in a given language, encodes it using the shared encoder $\mathcal{E}$, then reconstructs it by the decoder $\mathcal{D}_2$. In order to make the encoder truly learn the compositionality of its input words in a language independent manner, we propose to introduce random noise in the input sentences. Inspired by the denoising autoencoders~\citep{vincent2010stacked,hill2016learning} where the model is trained to reconstruct the original version of a corrupted input sentence, we alter the word order of $x$ on the output side by making random swaps between contiguous words. More concretely, for the reference sentence $x$ (which is also the input) whose length is $m$, we make $m/4$ random swaps of this kind. Denote the disordered sentence $x$ as $\tilde{x}$, the objective function can be defined as: 

\begin{equation}
\begin{aligned}
\mathcal{J}_D (\Theta_2) & =  \sum_{x \in D_x} P(\tilde{x}|x; \Theta_2)\\
& = \sum_{x \in D_x} \sum_{t = 1}^{m} {\rm log} P(\tilde{x}_t|\tilde{x}_{<t}, x; \Theta_2)
\end{aligned}
\end{equation}

In this way, the model needs to learn about the internal knowledge of the languages without the information of the correct word order. At the same time, by discouraging the model to rely too much on the word order of the input sequence, we can better account for the actual word order divergences across languages.

\subsection{Reinforcement Learning}
Different from translating the source language to the target language, our auto-encoder architecture is more like a subsidiary role, which aims to capture the main information of the text rather than transforming each exact word. Many words have synonyms which may distribute closely in the word embedding space but regarded as errors in the cross entropy objectives. To tackle this problem, inspired by previous works that leverage the technology of reinforcement learning~\cite{pan2018discourse,pan2019reinforced}, we use REINFORCE~\citep{williams1992simple} algorithm to maximize the expected reward, which is defined as:

\begin{equation}
\begin{aligned}
J_{RL}(\Theta_2) = \mathbb{E}_{x' \sim \pi_{\Theta_2} (x' | x)}[R(x', x)]
\end{aligned}
\end{equation}
where $\pi_{\Theta_2} (x' | x)$ is the previous action policy that reconstructs the sentence $x$, $x'$ is obtained by sampling from the predicted probability distribution $\pi (x' | x)$ from $\mathcal{D}_2$, and:

\begin{equation}
\begin{aligned}
R(x', x) = cos(x',x)
\end{aligned}
\end{equation}
is the reward function defined to use cosine function to measure the overlap of the embedding space between the predicted text and the input text. To this end, we maximize the expectation of the similarity of $x$ and the reconstructed $x'$ to ignore the grammatical structure and effectively pay more attention on the key information.

\begin{algorithm*}[t]
\caption{ \label{al1} Bi-Decoder Augmented Network} 
\KwIn{Training data sentence pairs ($D_{x}$, $D_{y}$); mini-batch size $m$; mixing ratio $\lambda_a, \lambda_d, \lambda_r$; the counting number $c$ = 0;}
\KwOut{The model's parameters $\theta_{e}, \theta_1, \theta_2$; } 	
\While{$\mathcal{D}_1$ not 90\% converge} 
{
	Ramdomly sample a mini-batch of $m$ sentence pairs $\{x^{(i)}, y^{(i)}\}_{m}$ from $D_{x}$ and $D_{y}$\;
	Updating $\mathcal{J}_1 (x^{(i)}, y^{(i)};\Theta_1))$ via supervised gradient\;
	\If{$c$ {\rm mod} $(\lambda_a + \lambda_d + \lambda_r) < \lambda_a$}
	{Optimizing $\mathcal{J}_2 (x^{(i)};\Theta_2)$ via supervised gradient;
	}
	\ElseIf{$c$ {\rm mod} $(\lambda_a + \lambda_d + \lambda_r) \geq \lambda_a + \lambda_d$}
	{Optimizing $\mathcal{J}_{RL} (x^{(i)};\Theta_2)$ via policy gradient;
	}
	\Else{Optimizing $\mathcal{J}_D (x^{(i)}, \tilde{x}^{(i)};\Theta_2)$ via supervised gradient;
	}
	$c=c+1$;
}
Fix $\theta_2$\;
\While{$\mathcal{D}_1$ not converge}
{
	Ramdomly sample a mini-batch of $m$ sentence pairs $\{x^{(i)}, y^{(i)}\}_{m}$ from $D_{x}$ and $D_{y}$\;
	Optimizing $\mathcal{J}_1 (x^{(i)}, y^{(i)};\Theta_1))$ via supervised gradient;
} 
\end{algorithm*}

\subsection{Multi-Objective Learning}
During training, we alternate the mini-batch optimization of the three objective functions, based on a tunable ``mixing ratio": $\lambda_a, \lambda_d, \lambda_r$, means that optimizing $\mathcal{J}_2$ for $\lambda_a$ mini-batches followed by optimizing $\mathcal{J}_D$ for $\lambda_d$ mini-batches, followed by optimizing $\mathcal{J}_{RL}$ for $\lambda_r$ mini-batches.

More importantly, instead of directly training the model until all of its parts converge, we first jointly train the parameters of the whole model to the decoder $\mathcal{D}_1$ 90\% convergence, and then fix $\theta_2$ (parameters of the $\mathcal{D}_2$) and train $\Theta_{1}$ using $\mathcal{J}_1$ until the model fully converges. This is because that the goal of the neural machine translation is to accurately transform the source language text to the target language text, so optimizing the decoder $\mathcal{D}_1$ is always the most important objective of the training procedure. By starting from a 90\% convergence baseline, we can avoid the model stucking in a local minimum. Based on that we propose our training algorithm in Algorithm \ref{al1}.

\section{Experiments}
\subsection{Implementation Details}
We evaluate our proposed algorithms and the baselines on three pairs of languages: English-to-German (En$\to$De), English-to-French (En$\to$Fr), and English-to-Vietnamese (En$\to$Vi). In detail, for En$\to$De and En$\to$Fr, we employ the standard filtered WMT’ 14, which is widely used in NMT evaluations~\citep{luong2015effective,bahdanau2015neural} and contains 1.9M and 2.0M training sentence pairs respectively\footnote{ \url{http://statmt.org/wmt14/translation-task.html}}. We test our models on \emph{newstest2014} in both directions for En$\to$De. For En$\to$Vi, we use IWSLT 2015, which is a smaller scale dataset and contains 133k training set and 1.2k testing set\footnote{ \url{https://nlp.stanford.edu/projects/nmt/data/}}.

We use the architecture from \citep{luong2015effective} as our baseline framework to construct our Bi-Decoder Augmented Network (BiDAN). We also employ the GNMT~\citep{wu2016google} attention to parallelize the decoder's computation. When training our NMT systems, we split the data into subword units using BPE~\citep{sennrich2016neural}. We train 4 layer LSTMs of 1024 units with bidirectional encoder, 4 layer unidirectional LSTMs of 1024 units for both $\mathcal{D}_1$ and $\mathcal{D}_2$, the embedding dimension is 1024. The mixing ratio $\lambda_a, \lambda_d, \lambda_r$ are set as 5:2:2. The model is trained with stochastic gradient descent with a learning rate that began at 1.0. We train for 680K steps; after 340K steps, we start halving learning rate every 34K step. For the baseline model, the batch size is set as 128 and for the BiDAN is 64, the dropout rate is 0.2. For the baseline and our BiDAN network, we use beam search with beam size 10 to generate sentences.

\begin{table*}[t]
\begin{center}
	\begin{tabular}{l|c|c|c|ccp{1cm}cp{1cm}cp{1cm}cp{1cm}}
		\toprule
		\multicolumn{1}{c|}{\multirow{2}{*}{\textbf{NMT Models}}} & \multicolumn{3}{c|}{\textbf{WMT14}} &   \multicolumn{1}{c}{\textbf{IWSLT15}} & \\ \cline{2-5} 
		\multicolumn{1}{c|}{}  & \textbf{En}$\to$\textbf{De} & 	\textbf{De$\to$En} & \textbf{En}$\to$\textbf{Fr} & 	\textbf{En$\to$Vi} & \\
		\midrule
		Baseline  & 22.6 & 26.8 & 32.3 & 24.9 \\ \hline
		Baseline + AD &  24.0 & 28.2  & 33.6 & 26.2\\
		Baseline + AD (Denoising) &  24.3 & 28.4 &  34.0 & 26.6\\ 
		Baseline + AD (RL) & 24.4 & 28.5 & 33.9 & 26.8 \\ \hline
		BiDAN (All modules converge) & 24.6 & 28.7 & 34.1 & 27.0 \\
		\textbf{BiDAN} & \textbf{24.7} & \textbf{28.9} & \textbf{34.2} & \textbf{27.1}   \\
		\bottomrule
	\end{tabular}
	\vspace{2mm}
\end{center}
\caption{\label{tab1} BLEU scores for NMT models on WMT14 English-German and English-French and IWSLT 2015 English-Vietnamese dataset. ``AD" denotes auxiliary decoder, ``RL" denotes reinforcement learning.}       
\end{table*}

\begin{table*}
\begin{center}
	\begin{tabular}{l|cc}
		\toprule
		\textbf{NMT Models} & \textbf{En}$\to$\textbf{De}& \textbf{De$\to$En}\\
		\midrule
		Transformer & 27.5 & 31.6 \\
		Transformer + AD & 28.1 & 32.0\\ 
		Transformer + AD (Denoising)& 28.0 & 31.9\\ 
		Transformer + AD (RL) & 28.1 & \textbf{32.1} \\ \hline
		\textbf{BiDAN (Transformer)}& \textbf{28.1} & 32.0\\  
		\bottomrule
	\end{tabular}
	\vspace{2mm}
\end{center}
\caption{\label{tab2} BLEU scores on WMT14 English-to-French for Transformer results.}       
\end{table*}

\subsection{Results}
As shown in the Table \ref{tab1}, our BiDAN model on all of the datasets performs much better compared to the baseline model. In the medium part, we conduct an ablation experiment to evaluate the individual contribution of each component of our model. At first, we only add the auxiliary decoder $\mathcal{D}_2$ to the baseline model, and the BLEU scores on all the test sets rise about 1.4 point, which shows the effectiveness of our bi-decoder architecture and the significance of the language-independent representation of the text. We then train our model with the process of denoising, in other words, we take out the objective function $\mathcal{J}_{RL}$ when optimizing our BiDAN model. We observe that the performance rises around 0.3 point, which proves that focusing on the internal structure of the languages is very helpful to the task. Finally, we use the reinforcement learning with the original loss to train our auxiliary decoder, which means we don't use the objective function $\mathcal{J}_D$. The results show that this leads to about 0.4 point improvement, which indicates that relaxing the grammatical limitation and capturing the keyword information is very useful in our bi-decoder architecture. Finally, instead of first jointly train the parameters of the whole model to the decoder $\mathcal{D}_1$ 90\% convergence and then fix $\theta_2$ (parameters of the $\mathcal{D}_2$) and train $\Theta_{1}$ using $\mathcal{J}_1$ until the model fully converges, we directly training the model until all of its parts converge. The results show that training from a well-trained model may induce the local minimum of the optimization.

\subsection{Transformer Ablation}
We also conduct experiments on the Transformer~\citep{vaswani2017attention}, which is another state-of-the-art architecture for NMT. We adopt the base model and setting in the official implementation\footnote{\url{https://github.com/tensorflow/models/tree/master/official/transformer}}. As depicted in Table \ref{tab2}, we can see that our auxiliary decoder improves the performance of the Transformer as well. However, the improvement from reinforcement learning is quite modest, while the denoising part even declines the scores. We conjecture that this is because the positional encoding of the Transformer reduces the order dependency capture of the model, thus counteracts the effects of these approaches.

\begin{table*}[t]
\begin{center}
	\begin{tabular}{l|cc}
		\toprule
		\textbf{Encoder Source} & \textbf{BLEU ($p_1$)} & \textbf{BLEU ($p_2$)}\\
		\midrule
		Random Encoder & 7.2 & 0.1 \\
		En$\to$De Encoder & 11.1 & 0.2\\ 
		En$\to$De Encoder (BiDAN)& 27.8 & 2.5\\ \hline
		En$\to$Fr Encoder (Original)& \textbf{62.4} & \textbf{39.1}\\  
		\bottomrule
	\end{tabular}
	\vspace{2mm}
\end{center}
\caption{\label{tab3} BLEU scores on WMT14 English-to-French for NMT models with different encoders.}       
\end{table*}

\subsection{Language Independence}
In the Table \ref{tab3}, we use encoders trained by different sources to evaluate the language-dependence extent of the high-level text representation of the models. We train the NMT model on WMT14 English-to-French dataset, and then replace the well-trained encoder with different sources at the testing time. Since the modified n-gram precision~\citep{papineni2002bleu} of BLEU $p_3$ and $p_4$ are all 0 except the original well-trained encoder, we use BLEU $p_1$ and $p_2$ to evlaute the difference among all the models. We first replace the encoder with random parameters and the results drop a lot, which is not surprising. Then we use the encoder trained on the WMT14 English-to-German dataset to test the performance of cross language performance. The results improve modestly compared to the random encoder but still have a huge gap to the original model. This indicates that the coarse cross language training doesn't work even though the languages are quite similar. Finally, we replace the original encoder with the encoder from our BiDAN framework, which is also trained on the WMT14 English-to-German dataset. Although the performance is still far away from the level which is achieved by the original encoder, it is much better than the previous two methods. We point out that the only difference among those models are the parameters of the encoder, which determine the text representation. Thus  the different performances demonstrate that our model provides a more language-dependent text representation, and this may also explain the improvement of our structure to the general NMT model.

\begin{figure}[t]
	\centering
	
	\includegraphics[scale=0.4]{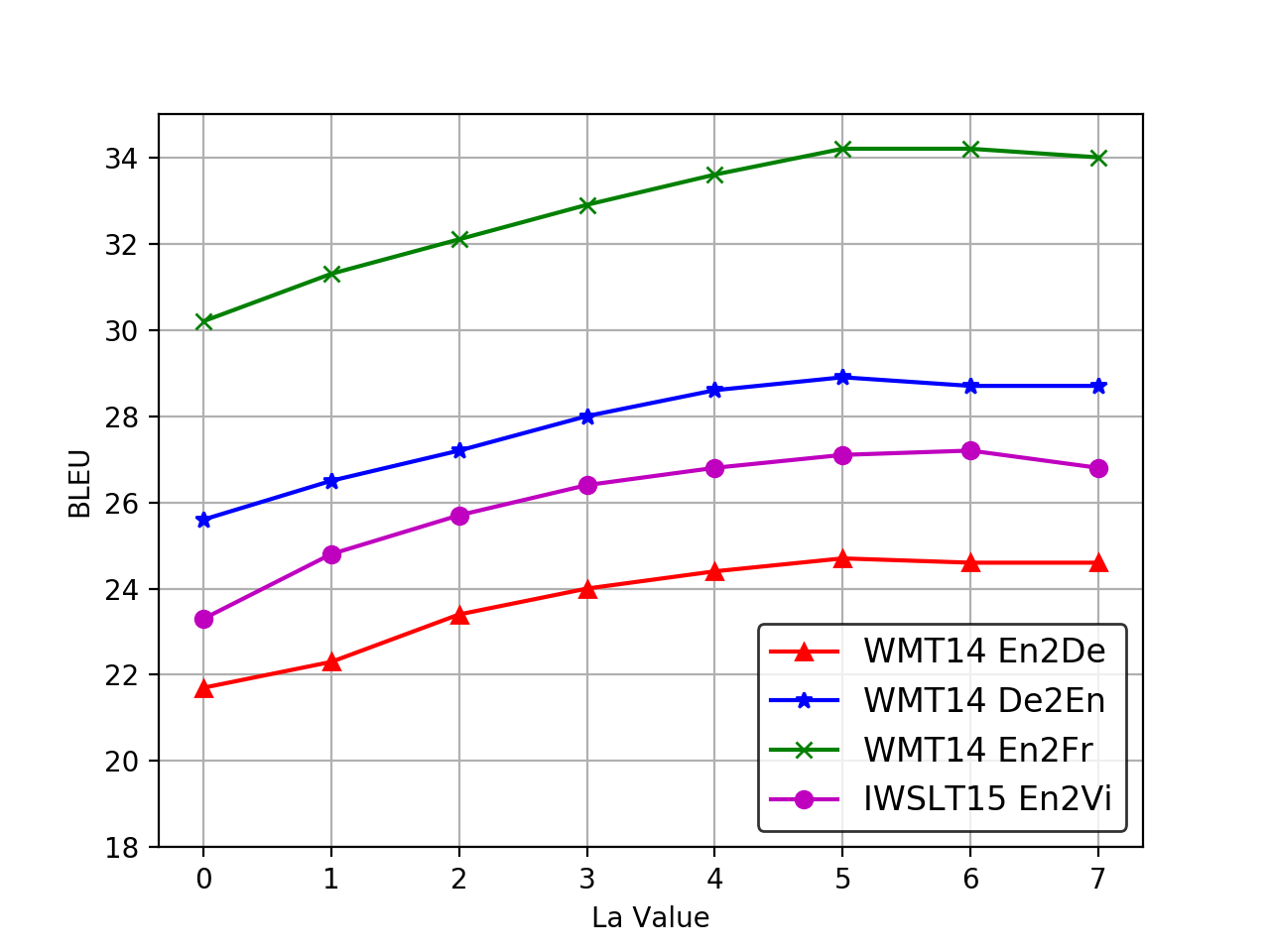}
	\caption{\label{fig2} Performance when the model is trained with different values of $\lambda_a$. }
\end{figure}

\begin{figure}[t]
	\centering
	\includegraphics[scale=0.4]{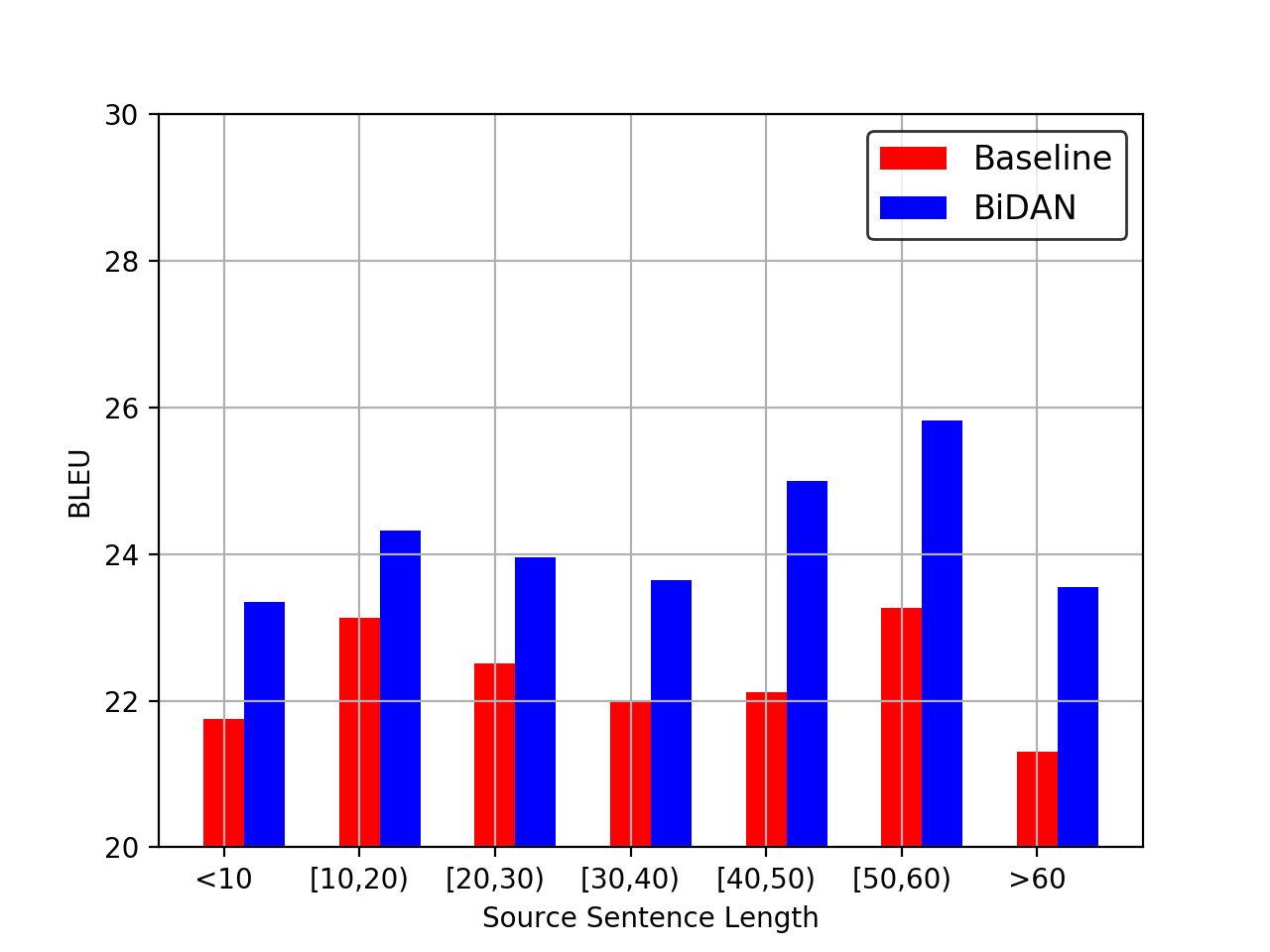}
	\caption{\label{fig3} Performance with different lengths of the source sentences on WMT14 English$\to$German.}
\end{figure}

\begin{table*}[t]
	\renewcommand\arraystretch{1.1}
	\begin{center}
		\begin{tabular}{p{2\columnwidth}}
			\toprule
			\textbf{Source: }Die T$\ddot{\rm {a}}$ter hatten Masken getragen und waren nicht erkannt worden.	\\
			\textbf{Reference: }The assailants had worn masks and had not been recognised.\\
			\textbf{Baseline: }The perpetrators had borne masks and were not recognized.\\ 
			\textbf{BiDAN: }The perpetrators had worn masks and had not been recognized.\\ \hline
			\textbf{Source: }Zu einem Grillfest bringt Proctor beispielsweise Auberginen, Champignons und auch ein paar Vegan w$\ddot{\rm {u}}$rste mit.\\
			\textbf{Reference: }For example, when attending a barbecue party, Proctor brings aubergines, mushrooms and also a few vegan sausages.\\
			\textbf{Baseline: }Proctor, for example, brings Aubergins, Champignons and a few Vegans.\\
			\textbf{BiDAN: }For example, Proctor brings aubergines, mushrooms and also a few vegan sausages at a barbecue.\\
			\bottomrule
		\end{tabular}
		\vspace{2mm}
	\end{center}
	\caption{\label{tab4} Examples of the generated sentences (\textbf{Baseline} and \textbf{BiDAN}) with the source and refenrence sentences (\textbf{Source} and \textbf{Reference}) from WMT14 German$\to$English dataset.}       
\end{table*}

\subsection{Visualization}
We present our model on the all the datasets with different values of $\lambda_a$ to show how the multi-objective learning affects the performance, as depicted in Figure \ref{fig2}. In other words, we keep $\lambda_d$ and $ \lambda_r$ as 2 and change the ratio of the original objective $\mathcal{J}_2$. As we can see, on all of the datasets the model drops sharply and even worse than the baseline model when we set $\lambda_a$ as 0, which means we only use the objective functions $\mathcal{J}_{D}$ and $\mathcal{J}_{RL}$ without the original one. This indicates that totally ignoring the grammatical structure of the input text is not helpful to the task. We also observe that the performance rises with the increase of $\lambda_a$ until 5 or 6. Afterwards, the results get worse when we raise the $\lambda_a$, which means the multi-objective learning can improve the performance. We didn't conduct more experiments with larger values of $\lambda_a$ because the ablation experiments show that the final results will converge at the BLEU values on the second row of the Table \ref{tab1}, which is about 0.7 point lower than the best performance.

Figure \ref{fig3} shows the performance with different lengths of the source sentences on WMT14 English$\to$German. As we can see, our method performs better than the baseline on all of the lengths. We also find that the extent of the improvement on long sentences is larger than that in short sentences. We conjecture this is because that our method is able to deeply understand the logical relations of the sentences via the language-independent semantic space, thus performs better on the long and complex sentences.

\subsection{Text Analysis}
In order to better understand the behavior of the proposed model, in the Table \ref{tab4}, we present some examples of the comparison among the generated sentences of our model and the baseline with the reference sentences from the WMT14 German$\to$English dataset.

In the first example, the reference translates the word ``\emph{getragen}" as ``\emph{worn}" and the word ``\emph{T$\ddot{a}$ter}" as ``\emph{assailants}". Both our method and the baseline model translate ``\emph{T$\ddot{a}$ter}" as ``\emph{perpetrators}", which is a synonym for the word ``\emph{assailants}". However, the baseline model translates the word ``\emph{getragen}" as ``\emph{borne}", which is the past participle form of the word ``\emph{bear}". Interestingly, the German word ``\emph{getragen}" is individually translated as ``carry", which is very similar to the word ``bear", but they are not suitable for the object ``mask" in English. This example proves that our model alleviates the weakness of the original unidirectional source-to-target architecture that is likely to induce the model to become a simple n-gram matching, which pays less attention to organizing natural languages.

For the second example, we can see that the translation given by the baseline model doesn't make sense from the perspective of semantic relation, while our model accurately translates the source text. It seems that the baseline is not good at recognizing the words when their first letters are capitalized. However, by generating a language-independent semantic space via the bi-decoder structure, our model can effectively understand the meanings of the sentences as the thinking process of human being.

\section{Related Works}
\label{sec4}
Research on language-independent representation of the text has attracted a lot of attention in recent times. Many significant works have been proposed to learn cross-lingual word embeddings~\citep{artetxe2016learning,smith2017offline}, which have a direct application in inherently cross-lingual tasks like machine translation \citep{Zou:EMNLP13}, cross-lingual entity linking~\citep{Tsai2016Cross}, and part-of-speech tagging~\citep{Gaddy2016Ten,pan2017memen}, \emph{etc}. However, very few works pay attention to the language-independent representations of the text for the higher layers of the neural networks, which may contain higher level semantic meanings and significantly affect the performance of the model. In this work, focusing on the representations generated from the encoder, we augment the natural language understanding of the NMT model by introducing an auxiliary decoder into the original encoder-decoder framework.

There have been several proposals to improve the source-to-target dependency of the sequence-to-sequence models~\citep{cheng2016semi,xia2017deliberation,tu2017neural,artetxe2018unsupervised,pan2018macnet,pan2019reinforced}. Closely related to our work, \citep{tu2017neural} proposed to reconstruct the input source sentence from the hidden layer of the output target sentence to ensure that the information in the source side is transformed to the target side as much as possible. However, their method use the decoder states as the input of the reconstructor and deeply relies on the hidden states of the decoder, which contributes less in learning the language-independent representations. \citep{artetxe2018unsupervised} use monolingual corpera to train a shared encoder based on the fixed cross-lingual embeddings. However, this work is proposed to learn an NMT system in a completely unsupervised manner that remove the need of parallel data, while our method still aims to augment the performance of source-to-target model based on supervised learning.

\section{Conclusions}
\label{sec5}

In this paper, we propose Bi-Decoder Augmented Network for the task of the neural machine translation. We design an architecture that contains two decoders based on the encoder-decoder model where one for generating the target language and another for reconstructing the source sentence. while being trained to transform the sequence into two languages, the model has the potential to generate a language-independent semantic space of the text. The experimental evaluation shows that our model achieves significant improvement on the baselines on several standard datasets. For the future works, we would like to explore whether we can extend our idea to the different target media such as images and text or the deeper level of the neural networks.

\section{Acknowledgement}
This work was supported in part by The National Key Research and Development Program of China (Grant Nos: 2018AAA0101400), in part by The National Nature Science Foundation of China (Grant Nos: 61936006), in part by the Alibaba-Zhejiang University Joint Institute of Frontier Technologies.

\section*{\refname}

\bibliographystyle{elsarticle-num}   	    
\bibliography{BiDAN_neurocomputing}

\begin{thebibliography}{10}
\expandafter\ifx\csname url\endcsname\relax
  \def\url#1{\texttt{#1}}\fi
\expandafter\ifx\csname urlprefix\endcsname\relax\def\urlprefix{URL }\fi
\expandafter\ifx\csname href\endcsname\relax
  \def\href#1#2{#2} \def\path#1{#1}\fi

\bibitem{luong2015effective}
T.~Luong, H.~Pham, C.~D. Manning, et~al., Effective approaches to
  attention-based neural machine translation, in: EMNLP, 2015, pp. 1412--1421.

\bibitem{he2017decoding}
D.~He, H.~Lu, Y.~Xia, T.~Qin, L.~Wang, T.~Liu, Decoding with value networks for
  neural machine translation, in: NIPS, 2017, pp. 178--187.

\bibitem{zhou2016deep}
J.~Zhou, Y.~Cao, X.~Wang, P.~Li, W.~Xu, Deep recurrent models with fast-forward
  connections for neural machine translation, in: Transactions of the
  Association of Computational Linguistics, 2016.

\bibitem{wu2016google}
Y.~Wu, M.~Schuster, Z.~Chen, Q.~V. Le, M.~Norouzi, W.~Macherey, M.~Krikun,
  Y.~Cao, Q.~Gao, K.~Macherey, et~al., Google's neural machine translation
  system: Bridging the gap between human and machine translation, in: arXiv
  preprint arXiv:1609.08144, 2016.

\bibitem{bahdanau2015neural}
D.~Bahdanau, K.~Cho, Y.~Bengio, et~al., Neural machine translation by jointly
  learning to align and translate, in: ICLR, 2015.

\bibitem{artetxe2016learning}
M.~Artetxe, G.~Labaka, E.~Agirre, Learning principled bilingual mappings of
  word embeddings while preserving monolingual invariance, in: EMNLP, 2016.

\bibitem{smith2017offline}
S.~L. Smith, D.~H. Turban, S.~Hamblin, N.~Y. Hammerla, Offline bilingual word
  vectors, orthogonal transformations and the inverted softmax, in: ICLR, 2017.

\bibitem{belinkov2017neural}
Y.~Belinkov, N.~Durrani, F.~Dalvi, H.~Sajjad, J.~Glass, What do neural machine
  translation models learn about morphology?, in: ACL, 2017.

\bibitem{guo2018soft}
H.~Guo, R.~Pasunuru, M.~Bansal, Soft layer-specific multi-task summarization
  with entailment and question generation, in: arXiv preprint arXiv:1805.11004,
  2018.

\bibitem{shi2016does}
X.~Shi, I.~Padhi, K.~Knight, Does string-based neural mt learn source syntax?,
  in: EMNLP, 2016, pp. 1526--1534.

\bibitem{robinson2012becoming}
D.~Robinson, Becoming a translator: An introduction to the theory and practice
  of translation, Routledge, 2012.

\bibitem{sutskever2014sequence}
I.~Sutskever, O.~Vinyals, Q.~V. Le, et~al., Sequence to sequence learning with
  neural networks, in: NIPS, 2014, pp. 3104--3112.

\bibitem{cho2014learning}
K.~Cho, B.~van Merrienboer, C.~Gulcehre, D.~Bahdanau, F.~Bougares, H.~Schwenk,
  Y.~Bengio, Learning phrase representations using rnn encoder--decoder for
  statistical machine translation, in: EMNLP, 2014, pp. 1724--1734.

\bibitem{vincent2010stacked}
P.~Vincent, H.~Larochelle, I.~Lajoie, Y.~Bengio, P.-A. Manzagol, Stacked
  denoising autoencoders: Learning useful representations in a deep network
  with a local denoising criterion, in: Journal of machine learning research,
  Vol.~11, 2010, pp. 3371--3408.

\bibitem{hill2016learning}
F.~Hill, K.~Cho, A.~Korhonen, Learning distributed representations of sentences
  from unlabelled data, in: Proceedings of NAACL-HLT, 2016, pp. 1367--1377.

\bibitem{pan2018discourse}
B.~Pan, Y.~Yang, Z.~Zhao, Y.~Zhuang, D.~Cai, X.~He, Discourse marker augmented
  network with reinforcement learning for natural language inference, in:
  Proceedings of the 56th Annual Meeting of the Association for Computational
  Linguistics (Volume 1: Long Papers), Vol.~1, 2018, pp. 989--999.

\bibitem{pan2019reinforced}
B.~Pan, H.~Li, Z.~Yao, D.~Cai, H.~Sun, Reinforced dynamic reasoning for
  conversational question generation, in: Proceedings of the 57th Annual
  Meeting of the Association for Computational Linguistics, 2019, pp.
  2114--2124.

\bibitem{williams1992simple}
R.~J. Williams, Simple statistical gradient-following algorithms for
  connectionist reinforcement learning, in: Machine learning, Vol.~8, Springer,
  1992, pp. 229--256.

\bibitem{sennrich2016neural}
R.~Sennrich, B.~Haddow, A.~Birch, et~al., Neural machine translation of rare
  words with subword units, in: ACL, 2016.

\bibitem{vaswani2017attention}
A.~Vaswani, N.~Shazeer, N.~Parmar, J.~Uszkoreit, L.~Jones, A.~N. Gomez,
  {\L}.~Kaiser, I.~Polosukhin, Attention is all you need, in: Advances in
  neural information processing systems, 2017, pp. 5998--6008.

\bibitem{papineni2002bleu}
K.~Papineni, S.~Roukos, T.~Ward, W.-J. Zhu, Bleu: a method for automatic
  evaluation of machine translation, in: ACL, Association for Computational
  Linguistics, 2002, pp. 311--318.

\bibitem{Zou:EMNLP13}
W.~Y. Zou, R.~Socher, D.~Cer, C.~D. Manning, Bilingual word embeddings for
  phrase-based machine translation, in: EMNLP, 2013.

\bibitem{Tsai2016Cross}
C.~T. Tsai, R.~Dan, Cross-lingual wikification using multilingual embeddings,
  in: Conference of the North American Chapter of the Association for
  Computational Linguistics: Human Language Technologies, 2016, pp. 589--598.

\bibitem{Gaddy2016Ten}
D.~M. Gaddy, Y.~Zhang, R.~Barzilay, T.~S. Jaakkola, Ten pairs to tag -
  multilingual pos tagging via coarse mapping between embeddings, in:
  Conference of the North American Chapter of the Association for Computational
  Linguistics: Human Language Technologies, 2016, pp. 1307--1317.

\bibitem{pan2017memen}
B.~Pan, H.~Li, Z.~Zhao, B.~Cao, D.~Cai, X.~He, Memen: Multi-layer embedding
  with memory networks for machine comprehension, arXiv preprint
  arXiv:1707.09098.

\bibitem{cheng2016semi}
Y.~Cheng, W.~Xu, Z.~He, W.~He, H.~Wu, M.~Sun, Y.~Liu, Semi-supervised learning
  for neural machine translation, in: Proceedings of the 54th Annual Meeting of
  the Association for Computational Linguistics, 2016.

\bibitem{xia2017deliberation}
Y.~Xia, F.~Tian, L.~Wu, J.~Lin, T.~Qin, N.~Yu, T.-Y. Liu, Deliberation
  networks: Sequence generation beyond one-pass decoding, in: Advances in
  Neural Information Processing Systems, 2017, pp. 1784--1794.

\bibitem{tu2017neural}
Z.~Tu, Y.~Liu, L.~Shang, X.~Liu, H.~Li, Neural machine translation with
  reconstruction., in: AAAI, 2017.

\bibitem{artetxe2018unsupervised}
M.~Artetxe, G.~Labaka, E.~Agirre, K.~Cho, Unsupervised neural machine
  translation, in: ICLR, 2018.

\bibitem{pan2018macnet}
B.~Pan, Y.~Yang, H.~Li, Z.~Zhao, Y.~Zhuang, D.~Cai, X.~He, Macnet: Transferring
  knowledge from machine comprehension to sequence-to-sequence models, in:
  Advances in Neural Information Processing Systems, 2018, pp. 6092--6102.

\end{thebibliography}
\end{document}